\documentclass[11pt]{article}

\usepackage[margin=1in]{geometry}

\usepackage{fontspec}
\usepackage{xeCJK}

\setCJKsansfont[
    BoldFont = FandolHei-Bold.otf
]{FandolHei-Regular.otf}

\setCJKmonofont{FandolFang-Regular.otf}

\newCJKfontfamily\cjkcodefont[
    Scale=0.80,
    AutoFakeBold=false
]{FandolSong-Regular.otf}

\newcommand{\cjktok}[1]{{\cjkcodefont\mdseries\upshape #1}}

\usepackage{algorithm}
\usepackage{algpseudocode}
\usepackage{amsmath, amssymb}
\usepackage{array}
\usepackage{booktabs}
\usepackage{caption}
\usepackage{csquotes}
\usepackage{graphicx}
\usepackage[hidelinks]{hyperref}
\hypersetup{
    pdftitle={Pruned BPE: Post-training Visibility Pruning and Token Reallocation for Byte Pair Encoding},
    pdfauthor={Kenny Shao},
    pdfsubject={Byte Pair Encoding and tokenizer vocabulary optimization},
    pdfkeywords={Byte Pair Encoding, BPE, tokenization, vocabulary pruning, subword tokenization}
}
\usepackage{multirow}
\usepackage{makecell}
\usepackage{placeins}
\usepackage{pgfplots}
\usepgfplotslibrary{groupplots}
\pgfplotsset{compat=1.18}

\pgfplotsset{
    standardbpe/.style={
        draw=blue!80!black,
        fill=blue!30,
        every node near coord/.append style={
            text=blue!80!black
        }
    },
    prunedbpe/.style={
        draw=purple!80!black,
        fill=purple!25,
        every node near coord/.append style={
            text=purple!80!black
        }
    }
}

\title{Pruned BPE: Post-training Visibility Pruning and Token Reallocation for Byte Pair Encoding}

\author{
Kenny Shao \\
Department of Computer Science \\
Florida International University \\
\texttt{kennyshao0919@gmail.com}
}

\date{}

\begin{document}

\maketitle

\begin{abstract}
Byte Pair Encoding (BPE) is widely used for subword tokenization, but standard BPE exposes every learned merge token to the downstream model, including tokens that mainly serve as intermediate construction units and rarely appear in the final encoded corpus. This paper proposes Pruned BPE, a post-training visibility-pruning and token-reallocation method that separates merge construction from model-visible vocabulary selection. After standard BPE training, tokens are evaluated by final exposure; low-exposure tokens are retained as internal-only merge nodes, while their model-visible vocabulary slots are reassigned to additional better-exposed candidates learned through resumed training, preserving the target model-visible vocabulary size. During encoding, internal-only tokens are recursively expanded into visible descendants before token IDs are returned, while the original BPE merge order is preserved. Experiments on two non-overlapping corpora dominated by English and Chinese, as well as on their combination, show that Pruned BPE consistently reduces encoded length relative to Standard BPE at the same training corpus, evaluation corpus, and model-visible vocabulary size. At a 40\% exposure threshold, Pruned BPE reduces encoded length by approximately 0.27\%--0.36\% on same-corpus evaluations under the same model-visible vocabulary budget. In an additional vocabulary-only evaluation using a shared exact minimum-token dynamic-programming encoder, Pruned BPE retains an advantage of approximately 0.23\%--0.31\%, providing evidence that the improvement arises from the more efficient composition of the model-visible vocabulary. Overall, these improvements represent a meaningful fraction of the approximately 1.5\%--3.8\% marginal reduction that would otherwise require adding another 2K tokens to the Standard BPE vocabulary. Qualitative analysis shows that internal-only tokens include reusable English word fragments, Chinese character and phrase components, partial UTF-8 byte sequences, and structured-text fragments. These results indicate that post-training visibility pruning can improve BPE vocabulary efficiency without increasing the number of tokens exposed to the language model.
\end{abstract}

\section{Introduction}
Byte Pair Encoding (BPE) was originally introduced by Gage in 1994 as a data-compression algorithm that iteratively replaces frequent adjacent byte pairs with new symbols~\cite{gage1994new}. It was later adapted as a subword-tokenization method for neural machine translation~\cite{sennrich2016neural} and became widely used in large-scale language modeling. Byte-level BPE, used in GPT-style tokenizers such as GPT-2~\cite{radford2019language}, is especially attractive for large language models (LLMs) because it provides a fixed vocabulary while retaining the ability to encode arbitrary Unicode text without an unknown-token issue.

BPE-style tokenizers remain common in modern LLM systems. For example, Meta's Llama 3 tokenizer uses \texttt{tiktoken} and BPE merge ranks~\cite{llama3tokenizer}; Alibaba's Qwen and Qwen3 models use byte-level BPE tokenization~\cite{qwen2023technical, qwen3technical}; and Mistral's v3 tokenizer is described as using BPE with SentencePiece~\cite{mistralv3tokenizer}. These examples show that BPE remains a practical tokenizer design, but they also motivate closer attention to how its limited vocabulary budget is allocated.

The allocation limitation comes from a mismatch between how BPE learns tokens and how the final vocabulary is used. During training, BPE repeatedly adds the most frequent adjacent pair as a new token. However, a token that is useful at an intermediate merge stage is not necessarily suitable as a final model-visible vocabulary entry. Some learned tokens mainly act as construction units for later, longer tokens. For example, fragments such as \texttt{vironment}, \texttt{ournal}, or \cjktok{朗普} may be useful on the path toward longer tokens such as \texttt{environment}, \texttt{environmental}, \texttt{journal}, \texttt{journalism}, \texttt{journalist}, or \cjktok{特朗普}, but they may be poor choices as final tokens if they rarely appear independently after all merges are applied. This problem is especially visible in byte-level tokenization. Since each Unicode character may be represented by multiple bytes, standard byte-level BPE may create intermediate byte combinations that are necessary for constructing complete characters, words, or multi-character expressions. For Chinese, Japanese, Korean, and other non-Latin scripts, some intermediate byte fragments may have little value as model-visible units even though they remain necessary for reconstructing the BPE merge tree. 

Exposing such low-exposure intermediate tokens to the language model may waste limited vocabulary capacity. Moreover, because these tokens are rarely emitted in the final encoded corpus, their embeddings may receive fewer effective training updates and may therefore be less well trained, potentially affecting downstream model performance.

This paper proposes \emph{Pruned BPE}, a post-training visibility-pruning and token-reallocation method for BPE tokenizers. The key idea is to separate \emph{merge construction} from \emph{vocabulary exposure}. BPE training is first allowed to build the merge tree normally. After training, learned tokens are evaluated by their \emph{final exposure}, defined as how often they remain visible in the fully encoded corpus. Tokens with insufficient final exposure are retained as internal-only construction tokens rather than exposed to the downstream language model. Their visible vocabulary slots are then reallocated to better candidate tokens learned by training beyond the target visible vocabulary size.

The method is therefore different from ordinary vocabulary pruning or trimming. It is not intended to shrink the final vocabulary. Instead, it preserves the target model-visible vocabulary size while changing which learned tokens are exposed. During encoding, standard BPE merging is preserved up to the final merged token sequence; any internal-only token is recursively expanded into visible child tokens before token IDs are returned. This design preserves the learned merge structure, allows visibility thresholds to be reapplied to a sufficiently large trained candidate merge structure, and frames the problem as vocabulary reallocation rather than simple token removal.

Overall, the contributions of this work are twofold:

First, this work introduces final exposure as a post-training criterion for separating BPE merge construction from model-visible vocabulary selection. It reallocates low-exposure token slots while preserving the target model-visible vocabulary size and the original merge order. The implementation also integrates token-ID remapping so that additional trained candidates can fill the target visible vocabulary size seamlessly and in a predictable manner. A vocabulary-only evaluation using a shared exact minimum-token objective further separates the effect of visible-vocabulary composition from native BPE merge-order encoding and shows that the resulting advantage persists.

Second, this work provides a practical reference implementation, including trainers and the tokenizer, which supports Unicode-aware pre-tokenization, token boundaries suitable for both natural languages and code-oriented text, CPU-parallel training, training-data sharding, checkpointing, and separate export of visible and internal vocabularies.

\section{Related Work}
BPE was first proposed by Gage as a compression algorithm based on replacing frequent adjacent byte pairs~\cite{gage1994new}. Sennrich et al. adapted BPE for neural machine translation, showing that subword units can represent rare words compositionally and reduce open-vocabulary problems~\cite{sennrich2016neural}. Later systems such as GPT-2, GPT-3, and RoBERTa adopted byte-level BPE or closely related variants for large-scale transformer models~\cite{radford2019language, brown2020language, liu2019roberta}. These works establish BPE as a practical tokenizer baseline, but they do not distinguish between tokens needed for merge construction and tokens that should remain model-visible after all merges are applied.

SentencePiece provides a language-independent tokenizer and detokenizer framework that can train subword models directly from raw text, including BPE and unigram language-model tokenization~\cite{kudo2018sentencepiece}. It is related to this work as a practical tokenizer-construction framework, but its main focus is training and detokenization infrastructure rather than post-training separation of internal merge tokens from model-visible vocabulary entries.

The following works are more directly related because they modify or reconsider how learned subword tokens are used after or during BPE-style training.

BPE-Dropout~\cite{provilkov2020bpe} modifies the deterministic BPE segmentation process by randomly dropping merge operations during training, producing multiple possible segmentations for the same word and improving robustness in some neural machine translation settings. This is different from Pruned BPE. BPE-Dropout changes the segmentation behavior during model training, while Pruned BPE preserves the standard BPE merge path and applies a post-training visibility decision to determine which learned tokens should be exposed to the model.

Scaffold-BPE~\cite{lian2025scaffold} is the most closely related work. It similarly identifies tokens that mainly serve as components of longer tokens, but it makes visibility decisions dynamically during BPE training. Pruned BPE instead preserves the standard BPE training path and applies post-training visibility pruning based on final exposure, allowing different visibility criteria to be applied to the same sufficiently large trained candidate merge structure. The main differences are summarized in Table~\ref{tab:scaffold_vs_pruned}.

\begin{table}[!htbp]
\centering
\scriptsize
\begin{tabular}{>{\raggedright\arraybackslash}p{0.24\linewidth}
                >{\raggedright\arraybackslash}p{0.34\linewidth}
                >{\raggedright\arraybackslash}p{0.34\linewidth}}
\toprule
\textbf{Aspect} & \textbf{Scaffold-BPE} & \textbf{Pruned BPE} \\
\midrule
When token visibility is decided
& During each BPE iteration when a new token is generated
& After a Standard BPE training stage; final visibility is assigned after resumed candidate training. \\
\midrule
Effect on visible-vocabulary construction path
& Online scaffold decisions can affect later visible-token membership
& Standard BPE merge path is preserved \\
\midrule
Required change to BPE training loop
& Yes
& The core pair-selection and merge operations are unchanged, but an outer resumed-training and dynamic-stopping procedure is added. \\
\midrule
Required change to BPE encoding implementation
& Yes
& Yes. Standard BPE merging is followed by recursive expansion of internal-only tokens, with merge ranks stored separately from exported IDs. \\
\midrule
Ability to change token visibility after training
& Limited; usually requires rerunning the online procedure
& Yes, provided that the trained candidate merge structure contains enough eligible tokens to fill the target model-visible vocabulary. \\
\midrule
Decision basis
& Dynamic frequency comparison against current queue-head frequency
& Final token exposure \\
\midrule
Main risk or tradeoff
& Early online decisions may be aggressive and may affect later visible vocabulary composition
& Requires training extra candidate tokens to refill the final visible vocabulary \\
\bottomrule
\end{tabular}
\caption{Comparison between Scaffold-BPE and Pruned BPE}
\label{tab:scaffold_vs_pruned}
\end{table}

Vocabulary-trimming methods also reconsider whether every learned subword should remain in the final vocabulary. Cognetta et al. study post-processing that replaces rare subwords with their components, but report no consistent performance improvement and possible substantial degradation~\cite{cognetta2024analysis}. Unlike simple trimming, Pruned BPE refills the released slots with additional trained candidates so that the target model-visible vocabulary size is preserved. The main differences are summarized in Table~\ref{tab:trimming_vs_pruned}.

\begin{table}[!htbp]
\centering
\scriptsize
\begin{tabular}{>{\raggedright\arraybackslash}p{0.24\linewidth}
                >{\raggedright\arraybackslash}p{0.34\linewidth}
                >{\raggedright\arraybackslash}p{0.34\linewidth}}
\toprule
\textbf{Aspect} & \textbf{Vocabulary Trimming} & \textbf{Pruned BPE} \\
\midrule
Main motivation and goal
& Remove rare or low-exposure subwords from the final vocabulary, mainly to reduce effective model vocabulary size and possibly improve robustness.
& Move low-exposure intermediate tokens out of the model-visible vocabulary and reallocate visible-token slots to other higher-exposure candidate tokens. \\
\midrule
Final visible vocabulary size and outcome
& Usually reduces the effective visible vocabulary size; decomposed tokens may also increase tokenized sequence length.
& Designed to preserve the target visible vocabulary size by refilling pruned slots with extra trained candidates; sequence length impact depends on the pruning criterion and replacement tokens. \\
\midrule
Treatment of removed or pruned tokens
& Removed from the final vocabulary and recursively decomposed when produced.
& Low-exposure tokens are retained internally, while visible slots are refilled from extra trained candidates. \\
\midrule
Token ID handling
& Often handled indirectly by downstream vocabulary construction; implementation details may depend on the toolkit.
& Explicit visible/internal vocabulary export with token-ID remapping. \\
\midrule
Required extra training budget
& No; starts from an already trained BPE vocabulary.
& Yes; resumed training continues until enough sufficiently exposed candidates are available to fill the target visible vocabulary. \\
\midrule
Main risk or tradeoff
& Can shrink the effective vocabulary and increase sequence length; prior experiments report no consistent performance improvement and possible degradation.
& Requires training extra candidate tokens and selecting an appropriate visibility criterion. \\
\midrule
Empirical evidence
& Prior NMT experiments found no consistent improvement and possible heavy degradation~\cite{cognetta2024analysis}.
& Evaluated in this work under fixed model-visible vocabulary sizes with resumed candidate training. \\
\bottomrule
\end{tabular}
\caption{Comparison between Vocabulary Trimming and Pruned BPE}
\label{tab:trimming_vs_pruned}
\end{table}

\FloatBarrier

\section{Proposed Method: Pruned BPE}

\subsection{Standard BPE Baseline}
The baseline in this work is standard byte-level BPE. Given a training corpus \(C\), each input string, or each chunk produced by the optional pretokenizer \(P\), is converted into a sequence of UTF-8 byte tokens. This initialization guarantees that any Unicode input can be represented without an unknown-token mechanism. The BPE training alphabet contains the 256 possible byte values. Any predefined special tokens occupy reserved positions in the final model-visible vocabulary but do not participate in merge learning.

During BPE training, the corpus is represented as a multiset of token sequences. At each step, the algorithm counts adjacent token pairs in the current corpus representation and selects the most frequent one. Let \(S\) denote the current multiset of token sequences, let \(f(x,y)\) denote the frequency of adjacent pair \((x,y)\) in \(S\), and let \(\mathcal{P}(S)\) denote the set of adjacent token pairs appearing in \(S\). The selected pair can be written as

\[
(a,b) = \arg\max \{ f(x,y) : (x,y) \in \mathcal{P}(S) \}.
\]

A new token $t$ is created to represent the concatenation of $a$ and $b$. The merge rule $(a,b) \rightarrow t$ is appended to the learned merge list, and all occurrences of the adjacent pair $(a,b)$ in the corpus representation are replaced by $t$. This process is repeated until the desired target vocabulary size is reached.

In standard BPE, every learned merge token is included in the final visible vocabulary. For the byte-level setting used in this work, the 256 base byte tokens and \(N_{\mathrm{reserved}}\) reserved special tokens occupy fixed positions in the model-visible vocabulary. Therefore, for a target model-visible vocabulary size \(N_{\mathrm{visible}}\), the required number of learned tokens is \(N_{\mathrm{required}} = N_{\mathrm{visible}} - 256 - N_{\mathrm{reserved}}\). Standard BPE performs \(N_{\mathrm{required}}\) successful merge operations and exposes every resulting learned token. The final vocabulary therefore contains the 256 base byte tokens, the \(N_{\mathrm{reserved}}\) reserved special tokens, and the
\(N_{\mathrm{required}}\) learned merge tokens.

Standard BPE therefore does not distinguish between learned tokens that remain frequent in the final encoded corpus and those that mainly serve as intermediate components of later merges. Pruned BPE retains the same merge process but changes how the final model-visible vocabulary is selected.

\subsection{Final Exposure and Visibility Pruning}
To distinguish these cases, the \emph{final exposure} of a token is defined as the number of times it appears in the corpus after the full BPE merge sequence has been applied. Let \(S^{*}\) denote the final encoded training corpus after all learned BPE merges have been performed. For a token \(t\), its final exposure is defined as

\[
E(t) = \sum_{s \in S^{*}} \operatorname{count}_{s}(t),
\]

where \(\operatorname{count}_{s}(t)\) is the number of occurrences of token \(t\) in the final token sequence \(s\). Final exposure is therefore different from the frequency of the pair that originally created the token. A token may be created from a highly frequent pair at an early training step, but later be absorbed into longer tokens and appear only rarely in the final encoded corpus.

Pruned BPE uses final exposure as a post-training visibility criterion. The merge tree is first trained normally using the standard BPE merge rule. The algorithm does not modify how adjacent pairs are counted, how the most frequent pair is selected, or how the corpus representation is updated during training. After training is complete, each learned token is evaluated according to its final exposure. Tokens whose exposure is greater than or equal to a threshold \(\tau\) are considered eligible for the model-visible vocabulary, while tokens whose exposure is below \(\tau\) are classified as internal-only. During final export, eligible learned tokens are admitted in original merge order until the target number of visible tokens is reached.

An internal-only token is not removed from the tokenizer. It remains in the merge table and can still be used as an intermediate construction unit during encoding. The difference is that it is never emitted as a model-visible token ID to the downstream language model. This separation allows the tokenizer to preserve the same merge structure learned by BPE while avoiding the use of model vocabulary slots for tokens that have little or no final exposure.

\subsection{Token Reallocation}
\label{subsec:token-reallocation}
If low-exposure tokens were simply removed from a Standard BPE vocabulary, the final model-visible vocabulary would become smaller than the target size. This would make comparison with Standard BPE unfair and could introduce the disadvantages associated with vocabulary trimming, including increased sequence length and possible downstream degradation~\cite{cognetta2024analysis}. Pruned BPE instead retains these tokens as internal-only merge nodes and assigns their released model-visible slots to sufficiently exposed candidates learned through resumed training. The method therefore performs vocabulary reallocation rather than vocabulary reduction and preserves the requested model-visible
vocabulary size.

A necessary implementation detail is that token reallocation changes exported token IDs but does not change the BPE merge order. During training, each learned token has an original merge rank determined by the order in which its merge was created. This rank must continue to control merge priority during encoding. The exported token ID, by contrast, identifies the token in the final visible or internal vocabulary. After pruning and reallocation, the exported token ID and original merge rank are not necessarily the same. For example, an internal-only token may be assigned an ID in the internal vocabulary while retaining an early merge rank. The tokenizer must therefore preserve both values so that reallocation does not alter the learned sequence of BPE merges.

In the reference implementation, token reallocation is performed in two stages. First, Standard BPE is trained to the target model-visible vocabulary size \(N_{\mathrm{visible}}\) with the minimum exposure threshold set to \(\tau=0\). Here, \(N_{\mathrm{visible}}\) denotes the complete vocabulary exposed to the downstream model, including the 256 base byte tokens and \(N_{\mathrm{reserved}}\) reserved special tokens. This produces the Standard BPE baseline and a checkpoint containing the merge structure learned up to the target size.

Second, the checkpoint is reloaded with the desired exposure threshold \(\tau>0\). Visibility analysis is applied to the learned tokens to measure the current number whose final exposure is at least \(\tau\). Training then resumes from the checkpoint to generate additional
candidate tokens. Rather than using a fixed extra-token budget, training continues until the number of sufficiently exposed learned tokens is large enough to fill all learned-token positions in the model-visible vocabulary:

\[
N_{\mathrm{eligible}}
\ge
N_{\mathrm{required}},
\qquad
N_{\mathrm{required}}
=
N_{\mathrm{visible}} - 256 - N_{\mathrm{reserved}}.
\]

Here, \(N_{\mathrm{eligible}}\) is the number of learned tokens whose final exposure is at least \(\tau\), and \(N_{\mathrm{required}}\) is the number of learned-token positions available in the model-visible vocabulary. The 256 base byte tokens and the \(N_{\mathrm{reserved}}\) reserved special tokens occupy the remaining fixed positions. A maximum training vocabulary size may be supplied as a safety bound, but the required training size is otherwise determined dynamically by this stopping condition.

After resumed training, final exposure is evaluated again on the resulting training corpus. Learned tokens are then scanned in original merge order. Low-exposure tokens encountered before the visible target
is filled are retained as internal-only tokens, while sufficiently exposed tokens are assigned to the model-visible vocabulary. Once the required number of visible learned tokens has been reached, export
stops and all remaining later candidates are discarded. Exported token IDs are assigned separately, while the original merge ranks of all retained learned tokens are preserved for encoding.

\subsection{Encoding with Internal-only Tokens}
During encoding, Pruned BPE applies all retained merge rules---those associated with visible or internal-only tokens---according to their original BPE merge ranks. Therefore, the initial output of the merge process may contain both visible tokens and internal-only tokens. Since internal-only tokens are not exposed to the language model, they must be expanded before the final token ID sequence is returned.

Each learned BPE token has two children corresponding to the pair that created it. If an encoded token is visible, its visible token ID is returned directly. If the token is internal-only, the tokenizer recursively expands it into its children. Each child is then checked in the same way: visible children are emitted, while internal-only children are expanded further. This process continues until all emitted tokens are visible tokens.

Formally, let \(\operatorname{child}(t)=(a,b)\) denote the two children of a learned token \(t\), and let \(\operatorname{id}_{\mathrm{export}}(t)\) denote the exported ID assigned to a visible token \(t\). The final emission function can be written recursively as

\[
\operatorname{emit}(t) =
\begin{cases}
[\operatorname{id}_{\mathrm{export}}(t)],
& \text{if } t \in V, \\[2pt]
\operatorname{emit}(a) \,\Vert\, \operatorname{emit}(b),
& \text{if } t \in I
  \text{ and } \operatorname{child}(t)=(a,b).
\end{cases}
\]

Here, \(V\) is the visible vocabulary, \(I\) is the internal-only vocabulary, and \(\Vert\) denotes sequence concatenation. Base byte tokens are always visible, so the recursion always terminates. As a result, Pruned BPE can use internal-only tokens during the merge process while guaranteeing that the final output sequence contains only model-visible token IDs.

This design preserves compatibility with ordinary downstream language-model training. The model only sees token IDs from the visible vocabulary. Internal-only tokens are implementation details of the tokenizer and are used only to support the merge structure needed to construct larger visible tokens.

\begin{algorithm}[!ht]
\caption{Pruned BPE Training with Post-training Visibility Pruning and Token Reallocation}
\label{alg:pruned-bpe}
\scriptsize
\begin{algorithmic}[1]

\Require Corpus \(C\); target model-visible vocabulary size
\(N_{\mathrm{visible}}\); reserved-token count \(N_{\mathrm{reserved}}\);
exposure threshold \(\tau\); optional maximum training size
\(N_{\mathrm{max}}\); pre-tokenizer \(P\), if used by the training pipeline

\Ensure Visible vocabulary \(V\); internal-only vocabulary \(I\);
retained merge table \(M_{\mathrm{export}}\); exported-ID mapping

\State Pretokenize \(C\) using \(P\), if applicable, and convert the chunks to UTF-8 byte sequences \(S\)
\State Initialize the 256 base byte tokens and the merge table \(M\gets\emptyset\)
\State \(N_{\mathrm{required}}\gets N_{\mathrm{visible}}-256-N_{\mathrm{reserved}}\)

\Statex \textbf{Stage 1: Standard BPE baseline}

\While{the number of learned tokens is less than \(N_{\mathrm{required}}\)}
    \State Perform one Standard BPE merge on \(S\)
    \State Record the merge rule and its original merge rank in \(M\)
\EndWhile

\State Save a checkpoint containing \(S\), \(M\), and the learned-token metadata
\State Export the corresponding Standard BPE tokenizer using \(\tau=0\)

\Statex \textbf{Stage 2: Resumed training with visibility pruning}

\State Reload the checkpoint and compute final exposure \(E(t)\)
\State \(N_{\mathrm{eligible}}\gets\) the number of learned tokens
\(t\) satisfying \(E(t)\geq\tau\)

\While{\(N_{\mathrm{eligible}}<N_{\mathrm{required}}\)}
    \If{\(N_{\mathrm{max}}\) is specified and the training vocabulary has reached \(N_{\mathrm{max}}\)}
    \State \textbf{terminate with an insufficient-candidates error}
\EndIf
    \State Perform a small batch of additional Standard BPE merges on \(S\), without exceeding \(N_{\mathrm{max}}\) when it is specified
    \State Record the new merge rules and their original merge ranks in \(M\)
    \State Recompute final exposure and update \(N_{\mathrm{eligible}}\)
\EndWhile

\Statex \textbf{Final visibility analysis and export}

\State Recompute final exposure \(E(t)\) on the resulting training corpus
\State Initialize \(V\) with the 256 base byte tokens and \(I\gets\emptyset\)

\For{each learned token \(t\) in original merge order}
    \If{the number of visible learned tokens has reached \(N_{\mathrm{required}}\)}
        \State \textbf{break}
    \ElsIf{\(E(t)<\tau\)}
        \State Add \(t\) to \(I\)
    \Else
        \State Add \(t\) to \(V\)
    \EndIf
\EndFor

\State Discard all remaining later candidate tokens
\State Append the \(N_{\mathrm{reserved}}\) special tokens to \(V\)
\State Retain in \(M_{\mathrm{export}}\) the merge rules for learned tokens retained in \(V\cup I\)
\State Assign exported IDs while preserving original merge ranks
\State \Return \(V\), \(I\), \(M_{\mathrm{export}}\), and the exported-ID mapping

\end{algorithmic}
\end{algorithm}

Algorithm~\ref{alg:pruned-bpe} summarizes the complete Pruned BPE training and export procedure. Here, \(N_{\mathrm{max}}\) bounds the trainable vocabulary consisting
of the 256 base byte tokens and the learned candidate tokens; reserved special tokens are excluded because they do not participate in BPE training. Pretokenization is not essential to the core method; the shared pretokenizer used by the reference implementation during training and encoding is described in Subsection~\ref{subsec:pretokenization}.

\FloatBarrier

\section{Experiments}

\subsection{Corpus Data}
Two tokenizer-training corpora were constructed for the experiments. They use disjoint document samples and do not share documents, although some source categories, such as Reddit and Chinese Wikipedia, are represented in both corpora. Both corpora were used during the development and testing of the proposed algorithm to verify the correctness of the trainer implementations and the tokenizer.

\paragraph{Corpus I.}
Corpus I contains approximately 640 MB of UTF-8 text. It consists primarily of English and Chinese text, with a small amount of source code and multilingual data. 
The English portion is approximately 430 MB and consists mainly of a locally collected sample from FineWeb-Edu, an educational English web-text dataset derived from the FineWeb/Common Crawl pipeline~\cite{penedo2024fineweb}, together with a sample of Reddit data~\cite{dewarim2017reddit}. The Chinese portion is approximately 204 MB and consists of conversational text collected from several Chinese social media platforms and text from Chinese Wikipedia pages, including both Simplified and Traditional Chinese. In addition, Corpus I contains a small amount of code-oriented text, including Java, Python, JavaScript, TypeScript, HTML, JSON, XML, and related formats, as well as a small multilingual component covering additional languages such as French, German, Portuguese, and Finnish.

\paragraph{Corpus II.}
Corpus II contains approximately 1 GB of UTF-8 text. It was designed to provide a larger and more balanced mixture of English, Chinese, multilingual text, and code-oriented text. The English portion includes a 360 MB subset sampled from CC-News and approximately 180 MB of randomly sampled Reddit data~\cite{dewarim2017reddit}. The CC-News subset was drawn from the Hugging Face \texttt{vblagoje/cc\_news} version of the Common Crawl News dataset~\cite{commoncrawl2016ccnews,vblagoje_ccnews}, which contains English-language news articles published between January 2017 and December 2019. To reduce ordering bias from source, crawl time, or topic clusters, the CC-News articles were randomly shuffled before sampling.

The Chinese portion of Corpus II includes a 382 MB subset sampled from THUCNews and approximately 30 MB of text from Chinese Wikipedia pages, including both Simplified and Traditional Chinese. THUCNews is a Chinese news text classification corpus released with the THUCTC project by the Natural Language Processing Laboratory of Tsinghua University~\cite{thuctc_thucnews}, and was generated from Sina News RSS subscription data from 2005 to 2011. The selected subset covers 14 news categories: sports, entertainment, home, lottery, real estate, education, fashion, politics, horoscopes, gaming, society, technology, stocks, and finance. To avoid category imbalance caused by directory ordering, THUCNews was sampled separately by category, with each category limited to the same maximum size before being added to the tokenizer-training corpus.

Corpus II also contains a small amount of code-oriented text, including Java, Python, JavaScript, TypeScript, C++, Markdown, and HTML. In addition, it includes an 84 MB multilingual corpus covering 42 languages other than English and Chinese, with approximately equal amounts of text for each language. These languages include Arabic, French, Spanish, Portuguese, Russian, Japanese, Korean, and many others.

\paragraph{Preprocessing.}
All text sources were lightly cleaned before tokenizer training. The preprocessing removed common boilerplate content, social-media sharing widgets or sharing blocks, duplicated adjacent lines, visible encoding artifacts, and website-specific templates or navigation texts. The goal of this cleaning step was not to heavily normalize the text, but to remove obvious non-content material that could distort token-frequency statistics. The corpora were otherwise kept as mixed natural UTF-8 text so that the tokenizer would be trained on realistic English, Chinese, multilingual, and code-oriented inputs.

\paragraph{Training and evaluation use.}
No separate held-out split was created. Same-corpus evaluations use the tokenizer-training corpus itself, while the additional cross-corpus evaluations are described in
Subsection~\ref{subsec:experimental-setup}.

\paragraph{Code and data availability.}
The Python and Cython implementations, the corpus data used in the experiments, and all trained tokenizer files are publicly available in the project repository~\cite{shao2026prunedbpe}. Each Standard BPE and Pruned BPE configuration includes a \texttt{vocab.txt} file. Pruned BPE configurations additionally include an \texttt{inter\_vocab.txt} file containing the internal-only vocabulary.

\subsection{Pretokenization}
\label{subsec:pretokenization}
Before applying BPE training or tokenization, input text is divided into boundary-aware text chunks using a lightweight pretokenization step. This step is Unicode-aware and is intended to preserve natural boundaries between scripts, numbers, punctuation, whitespace, and code-like substrings. Pretokenization is used as an implementation choice to make the tokenizer behavior more stable on mixed English, Chinese, multilingual, and code-oriented text. It is not the main contribution of this work.

The pretokenizer preserves the original input exactly: concatenating all produced chunks reconstructs the original string without loss. Ordinary single spaces are attached to the following word-like chunk when possible, following the common convention used by GPT-style BPE tokenizers in which leading spaces are treated as part of the following token. However, tabs, newlines, indentation, and runs of multiple spaces are kept as separate chunks so that formatting-sensitive text, especially source code, is not normalized or destroyed.

Characters are first classified into broad Unicode-aware categories, including whitespace, Latin word characters, digits, non-Latin Unicode word characters, punctuation, symbols, and other characters. Latin word spans include ASCII letters, accented Latin letters, digits, and underscores, allowing terms such as \texttt{abc123}, \texttt{GPT4}, \texttt{IPv6}, \texttt{user\_id}, \texttt{don't}, \texttt{C'était}, \texttt{déjà}, and \texttt{Montréal} to remain within a single pretokenized chunk when appropriate. The implementation also keeps common code- and technical-expression patterns such as \texttt{GPT-4}, \texttt{COVID-19}, \texttt{Node.js}, \texttt{file.java}, \texttt{C++}, and \texttt{C\#} as compact chunks when their internal punctuation functions as part of the expression rather than as a structural boundary.

Digit-first spans are handled separately. Numeric expressions such as \texttt{3.14} and \texttt{10-20} are kept together, while mixed number-first expressions such as \texttt{123abc} are split at the transition from number to letters. This rule also helps avoid merging Chinese date-like strings into large mixed chunks. For example, a string such as \cjktok{1949年6月24日} is segmented into alternating numeric and Chinese word chunks.

Non-Latin Unicode word spans, including Chinese, Japanese, Korean, Greek, Cyrillic, and other scripts, are grouped separately from Latin words and digits. This prevents cross-script chunks such as Chinese-plus-English or Japanese-plus-number sequences from being treated as a single initial unit. For example, Chinese text followed by an English abbreviation is split at the script boundary, and Japanese or Korean text followed by digits is split at the transition to the numeric span.

The pretokenizer also treats structural punctuation as boundaries. Punctuation and symbols are generally emitted as separate chunks, while repeated identical punctuation or symbol characters are grouped together, such as repeated periods, dashes, equal signs, or closing parentheses. A small number of code-oriented structural cases are handled explicitly; for example, the opener of an HTML or XML closing tag is preserved as \texttt{</}, so that \texttt{</div>} is pretokenized into \texttt{</}, \texttt{div}, and \texttt{>}. File-extension-like chunks such as \texttt{.js}, \texttt{.java}, \texttt{.py}, \texttt{.html}, and \texttt{.json} are also kept together when they occur in appropriate contexts.

This pretokenization step is therefore designed to reduce undesirable merges across obvious linguistic, numeric, punctuation, and structural boundaries while still preserving useful word-like and code-like units. It does not change the BPE merge algorithm itself; it only defines the initial text chunks over which BPE is applied. 
To faithfully compare Standard BPE with the proposed Pruned BPE method, the same pretokenization procedure is applied to both methods during training and tokenization in the experiments described below.

\subsection{Experimental Setup}
\label{subsec:experimental-setup}
Using Corpus I, Corpus II, and the combined Corpus I + Corpus II described above, three corpus-specific tokenizer training configurations were evaluated. All three configurations followed the same implementation and experimental procedure, differing only in the corpus used for tokenizer training.

The target vocabulary sizes were selected using a marginal encoded-length reduction heuristic. During tokenizer training, multiple checkpoints were saved at increasing vocabulary sizes. The Standard BPE tokenizer constructed from each checkpoint was used to encode the corresponding training corpus, and the marginal reduction in encoded length was measured between adjacent checkpoints. 
For two adjacent candidate values \(N_i<N_{i+1}\) of the model-visible vocabulary size \(N_{\mathrm{visible}}\), let \(L_{\mathrm{std}}(N_i)\) and \(L_{\mathrm{std}}(N_{i+1})\) denote the numbers of tokens produced when the same corpus is encoded by the corresponding Standard BPE tokenizers. The marginal encoded length reduction is defined as:

\begin{equation}
R_{\mathrm{std}}(N_i,N_{i+1})
=
\frac{L_{\mathrm{std}}(N_i)-L_{\mathrm{std}}(N_{i+1})}
     {L_{\mathrm{std}}(N_i)}
\times 100\%.
\label{eq:standard-marginal-reduction}
\end{equation}

For each training corpus, four consecutive candidate target vocabulary sizes at 2K intervals were selected from the region immediately before the marginal reduction curve began to flatten. This region continued to provide meaningful compression benefits while avoiding vocabulary sizes at which additional merges exhibited clear diminishing returns. To calculate the marginal reduction associated with the largest selected target size, one additional Standard BPE checkpoint 2K above that size was retained. This additional checkpoint was used only for the marginal-reduction calculation and was not included as a Pruned BPE evaluation target. This selection procedure was used only to identify reasonable experimental vocabulary budgets and is not part of the proposed pruning method. The main evaluation compares Standard BPE and Pruned BPE at the same selected target vocabulary sizes.

\begin{table}[!htb]
\centering
\scriptsize
\setlength{\tabcolsep}{3pt}
\renewcommand{\arraystretch}{1.08}

\begin{tabular}{@{}llcrrrrrrr@{}}
\toprule
\multirow{2}{*}{\makecell{Training\\corpus}} &
\multirow{2}{*}{\makecell{Evaluation\\corpus}} &
\multirow{2}{*}{\makecell{Model-visible\\vocab. size}} &
\multirow{2}{*}{\makecell{Standard BPE\\length (M)}} &
\multicolumn{2}{c}{Pruned BPE (20\%)} &
\multicolumn{2}{c}{Pruned BPE (30\%)} &
\multicolumn{2}{c}{Pruned BPE (40\%)} \\
\cmidrule(lr){5-6}
\cmidrule(lr){7-8}
\cmidrule(lr){9-10}
& & & &
Length (M) & $\Delta_{20}$ (\%) &
Length (M) & $\Delta_{30}$ (\%) &
Length (M) & $\Delta_{40}$ (\%) \\
\midrule

\multirow{8}{*}{Corpus I}
& \multirow{4}{*}{Corpus I}
& 8K  & 194.009 & 193.693 & 0.16 & 193.552 & 0.24 & 193.403 & 0.31 \\
& & 10K & 186.742 & 186.461 & 0.15 & 186.321 & 0.23 & 186.199 & 0.29 \\
& & 12K & 181.310 & 181.053 & 0.14 & 180.935 & 0.21 & 180.803 & 0.28 \\
& & 14K & 177.056 & 176.793 & 0.15 & 176.666 & 0.22 & 176.566 & 0.28 \\
\cmidrule(lr){2-10}
& \multirow{4}{*}{Corpus II}
& 8K  & 372.997 & 372.555 & 0.12 & 372.238 & 0.20 & 371.942 & 0.28 \\
& & 10K & 360.776 & 360.306 & 0.13 & 360.069 & 0.20 & 359.732 & 0.29 \\
& & 12K & 351.264 & 350.865 & 0.11 & 350.646 & 0.18 & 350.420 & 0.24 \\
& & 14K & 343.806 & 343.292 & 0.15 & 343.122 & 0.20 & 342.934 & 0.25 \\
\midrule

\multirow{8}{*}{Corpus II}
& \multirow{4}{*}{Corpus I}
& 10K & 197.549 & 197.311 & 0.12 & 197.220 & 0.17 & 197.119 & 0.22 \\
& & 12K & 191.933 & 191.624 & 0.16 & 191.491 & 0.23 & 191.417 & 0.27 \\
& & 14K & 187.445 & 187.183 & 0.14 & 187.058 & 0.21 & 187.004 & 0.24 \\
& & 16K & 183.931 & 183.604 & 0.18 & 183.460 & 0.26 & 183.342 & 0.32 \\
\cmidrule(lr){2-10}
& \multirow{4}{*}{Corpus II}
& 10K & 319.106 & 318.580 & 0.16 & 318.297 & 0.25 & 318.048 & 0.33 \\
& & 12K & 308.820 & 308.271 & 0.18 & 308.019 & 0.26 & 307.787 & 0.33 \\
& & 14K & 300.691 & 300.148 & 0.18 & 299.874 & 0.27 & 299.700 & 0.33 \\
& & 16K & 294.086 & 293.492 & 0.20 & 293.237 & 0.29 & 293.014 & 0.36 \\
\midrule

\multirow{12}{*}{\makecell{Corpus I\\+\\Corpus II}}
& \multirow{4}{*}{Corpus I}
& 12K & 186.137 & 185.905 & 0.12 & 185.754 & 0.21 & 185.637 & 0.27 \\
& & 14K & 181.753 & 181.521 & 0.13 & 181.380 & 0.21 & 181.284 & 0.26 \\
& & 16K & 178.266 & 178.036 & 0.13 & 177.940 & 0.18 & 177.836 & 0.24 \\
& & 18K & 175.320 & 175.101 & 0.13 & 174.995 & 0.19 & 174.927 & 0.22 \\
\cmidrule(lr){2-10}
& \multirow{4}{*}{Corpus II}
& 12K & 310.885 & 310.493 & 0.13 & 310.237 & 0.21 & 310.034 & 0.27 \\
& & 14K & 302.763 & 302.333 & 0.14 & 302.119 & 0.21 & 301.879 & 0.29 \\
& & 16K & 296.077 & 295.601 & 0.16 & 295.410 & 0.23 & 295.185 & 0.30 \\
& & 18K & 290.537 & 290.049 & 0.17 & 289.856 & 0.23 & 289.666 & 0.30 \\
\cmidrule(lr){2-10}
& \multirow{4}{*}{\makecell{Corpus I\\+\\Corpus II}}
& 12K & 497.022 & 496.399 & 0.13 & 495.992 & 0.21 & 495.671 & 0.27 \\
& & 14K & 484.516 & 483.854 & 0.14 & 483.500 & 0.21 & 483.164 & 0.28 \\
& & 16K & 474.343 & 473.637 & 0.15 & 473.350 & 0.21 & 473.022 & 0.28 \\
& & 18K & 465.857 & 465.149 & 0.15 & 464.851 & 0.22 & 464.593 & 0.27 \\
\bottomrule
\end{tabular}
\caption{Encoded lengths produced by Standard BPE and Pruned BPE under different exposure thresholds.}
\label{tab:encoded-length-reduction}
\end{table}

For each selected target size, the Standard BPE checkpoint produced at that size was used both to export the baseline tokenizer and to initialize three Pruned BPE variants. Let \(t_b\) denote the last learned token produced during Stage~1, that is, the token created by the \(N_{\mathrm{required}}\)-th successful merge, and let \(F(t_b)\) denote the BPE-training frequency recorded when \(t_b\) is created. This quantity is distinct from the final exposure \(E(t)\) recomputed after training and used by Algorithm~\ref{alg:pruned-bpe} to determine token visibility. The minimum exposure thresholds were defined as

\begin{equation}
\tau_r
=
\left\lceil r \cdot F(t_b) \right\rceil,
\qquad
r \in \{0.20, 0.30, 0.40\}.
\label{eq:min-exposure-threshold}
\end{equation}

For each value of \(r\), Algorithm~\ref{alg:pruned-bpe} is executed with its exposure-threshold input set to \(\tau=\tau_r\). Thus, the 20\%, 30\%, and 40\% settings correspond to minimum exposure counts equal to 20\%, 30\%, and 40\%, respectively, of the BPE-training frequency of the Stage~1 boundary token, rounded upward to the nearest integer. For each threshold, training was resumed from the corresponding checkpoint using the dynamic stopping and final-export procedure defined in Subsection~\ref{subsec:token-reallocation}.

In the experiments reported here, \(N_{\mathrm{reserved}}=4\). The four reserved special tokens were \texttt{<\_EOS\_>}, \texttt{<\_SOS\_>}, \texttt{<\_SEP\_>}, and \texttt{<\_PAD\_>}. They were appended to the model-visible vocabulary during export and were excluded from BPE training and visibility pruning. The model-visible and internal-only vocabularies
were exported as \texttt{vocab.txt} and \texttt{inter\_vocab.txt}, respectively.

The resulting Standard BPE tokenizer and its three Pruned BPE variants were evaluated on both the tokenizer training corpus and one or more additional evaluation corpora. For tokenizers trained on a single corpus, the other corpus served as a cross-corpus evaluation set. Tokenizers trained on the combined corpus were evaluated on the combined corpus and on each component corpus separately.

For \(p\in\{20,30,40\}\), let \(L_{\mathrm{pruned},p}(N)\) denote the encoded length produced by the Pruned BPE tokenizer using the \(p\%\) threshold setting at model-visible vocabulary size \(N\). Its reduction relative to the corresponding Standard BPE tokenizer is

\begin{equation}
\Delta_p(N)
=
\frac{
L_{\mathrm{std}}(N)-L_{\mathrm{pruned},p}(N)
}{
L_{\mathrm{std}}(N)
}
\times 100\%.
\label{eq:pruned-bpe-reduction}
\end{equation}

Each comparison uses the same training corpus, evaluation corpus, and model-visible vocabulary size.

Encoded length is the total number of tokens produced when a corpus is encoded by a given tokenizer. Two reduction measures are used. Equation~\eqref{eq:standard-marginal-reduction} measures the marginal benefit of increasing the Standard BPE vocabulary between adjacent target sizes. Equation~\eqref{eq:pruned-bpe-reduction} measures the additional reduction obtained by Pruned BPE relative to Standard BPE at the same training corpus, evaluation corpus, and model-visible vocabulary size.

For threshold setting \(p\), let \(N_{\mathrm{internal},p}\) denote the number of retained internal-only tokens. The internal-only share is defined as

\[
\mathrm{Share}_p
=
\frac{N_{\mathrm{internal},p}}
     {N_{\mathrm{required}}}
\times 100\%.
\]

Thus, the share is measured relative to the number of learned-token positions in the model-visible vocabulary.

Table~\ref{tab:encoded-length-reduction} reports the encoded lengths and corresponding Pruned BPE reduction percentages, with encoded lengths expressed in millions of tokens. All reduction percentages were calculated from the original integer token counts before the encoded lengths were rounded for presentation in millions. Table~\ref{tab:internal-only-token-stats} separately reports the counts and shares of tokens classified as internal-only for each Pruned BPE tokenizer.

\begin{table}[!htbp]
\centering
\small
\setlength{\tabcolsep}{4pt}
\renewcommand{\arraystretch}{1.15}

\begin{tabular}{@{}lcrrrrrr@{}}
\toprule
\multirow{2}{*}{\makecell{Training\\corpus}} &
\multirow{2}{*}{\makecell{Model-visible\\vocabulary size}} &
\multicolumn{2}{c}{20\% threshold} &
\multicolumn{2}{c}{30\% threshold} &
\multicolumn{2}{c}{40\% threshold} \\
\cmidrule(lr){3-4}
\cmidrule(lr){5-6}
\cmidrule(lr){7-8}
& & Count & Share (\%) & Count & Share (\%) & Count & Share (\%) \\
\midrule

\multirow{4}{*}{Corpus I}
 & 8K  &  84 & 1.09 & 131 & 1.69 & 189 & 2.44 \\
 & 10K & 101 & 1.04 & 162 & 1.66 & 227 & 2.33 \\
 & 12K & 121 & 1.03 & 192 & 1.64 & 278 & 2.37 \\
 & 14K & 154 & 1.12 & 246 & 1.79 & 335 & 2.44 \\
\midrule

\multirow{4}{*}{Corpus II}
 & 10K & 101 & 1.04 & 167 & 1.71 & 238 & 2.44 \\
 & 12K & 136 & 1.16 & 212 & 1.81 & 294 & 2.50 \\
 & 14K & 168 & 1.22 & 272 & 1.98 & 350 & 2.55 \\
 & 16K & 221 & 1.40 & 338 & 2.15 & 458 & 2.91 \\
\midrule

\multirow{4}{*}{\makecell{Corpus I\\+\\Corpus II}}
 & 12K & 100 & 0.85 & 179 & 1.52 & 256 & 2.18 \\
 & 14K & 134 & 0.98 & 219 & 1.59 & 315 & 2.29 \\
 & 16K & 170 & 1.08 & 258 & 1.64 & 371 & 2.36 \\
 & 18K & 207 & 1.17 & 311 & 1.75 & 419 & 2.36 \\
\bottomrule
\end{tabular}
\caption{Counts and shares of internal-only tokens under different exposure thresholds.}
\label{tab:internal-only-token-stats}
\end{table}

\subsection{Effect of the Exposure Threshold}
Together, Tables~\ref{tab:encoded-length-reduction} and \ref{tab:internal-only-token-stats} show the effect of increasing the exposure threshold from 20\% to 40\%. For a fixed training corpus and model-visible vocabulary size, the internal-only token sets are nested: every token classified as internal-only at a lower threshold remains internal-only at a higher threshold, while additional low-exposure tokens may also be classified as internal-only. Consequently, the number of internal-only tokens is non-decreasing as the exposure threshold rises, while the internal-only token share also generally increases. With respect to the other experimental dimension, as the model-visible vocabulary size grows, the number of internal-only tokens is likewise non-decreasing, while the share generally increases but may remain stable or fluctuate slightly because the total vocabulary size grows simultaneously. One possible explanation for this overall upward trend is that continued training creates longer tokens that absorb occurrences previously represented by shorter intermediate tokens, thereby reducing the final exposure of those intermediate tokens.

The encoded-length results show a corresponding improvement. For every reported combination of training corpus, evaluation corpus, and model-visible vocabulary size, a higher exposure threshold produces a greater reduction in encoded length relative to the corresponding Standard BPE tokenizer. Within the tested threshold range, pruning more low-exposure tokens therefore allows more visible vocabulary slots to be reallocated and consistently improves encoded-length efficiency.

\subsection{Comparison with Standard BPE}
\label{subsec:comp_std_bpe}

\begin{figure}[!htbp]
\centering

\begin{tikzpicture}
\begin{groupplot}[
    group style={
        group size=3 by 1,
        horizontal sep=0.8cm,
        ylabels at=edge left,
        yticklabels at=edge left
    },
    width=0.32\textwidth,
    height=0.34\textwidth,
    ybar,
    /pgf/bar width=5pt,
    ymin=0,
    ymax=4.5,
    ytick={0,1,2,3,4},
    ylabel={Encoded-length reduction (\%)},
    xlabel={Model-visible vocabulary size \(N_{\mathrm{visible}}\)},
    symbolic x coords={8K,10K,12K,14K,16K,18K},
    xtick=data,
    enlarge x limits=0.18,
    tick label style={font=\scriptsize},
    label style={font=\scriptsize},
    title style={font=\small},
    legend style={
        font=\scriptsize,
        draw=none,
        at={(0.5,-0.30)},
        anchor=north,
        legend columns=2,
        legend cell align={left},
        /tikz/every even column/.append style={column sep=2em}
    },
    nodes near coords,
    nodes near coords style={
        font=\tiny,
        rotate=90,
        anchor=west,
        /pgf/number format/fixed,
        /pgf/number format/precision=2,
        /pgf/number format/fixed zerofill
    },
    every axis plot/.append style={
        fill opacity=0.75,
        draw opacity=1
    }
]

\nextgroupplot[
    title={(a) Corpus I},
    symbolic x coords={8K,10K,12K,14K},
    xtick=data
]

\addplot[standardbpe] coordinates {
    (8K,3.75)
    (10K,2.91)
    (12K,2.35)
    (14K,1.95)
};

\addplot[prunedbpe] coordinates {
    (8K,0.31)
    (10K,0.29)
    (12K,0.28)
    (14K,0.28)
};

\nextgroupplot[
    title={(b) Corpus II},
    symbolic x coords={10K,12K,14K,16K},
    xtick=data
]

\addplot[standardbpe] coordinates {
    (10K,3.22)
    (12K,2.63)
    (14K,2.20)
    (16K,1.87)
};

\addplot[prunedbpe] coordinates {
    (10K,0.33)
    (12K,0.33)
    (14K,0.33)
    (16K,0.36)
};

\legend{
    Standard BPE:
    \(N_{\mathrm{visible}}\!\rightarrow\!
    N_{\mathrm{visible}}+2\mathrm{K}\),
    Pruned BPE: Fixed \(N_{\mathrm{visible}}\)
}

\nextgroupplot[
    title={(c) Corpus I + Corpus II},
    symbolic x coords={12K,14K,16K,18K},
    xtick=data
]

\addplot[standardbpe] coordinates {
    (12K,2.52)
    (14K,2.10)
    (16K,1.79)
    (18K,1.54)
};

\addplot[prunedbpe] coordinates {
    (12K,0.27)
    (14K,0.28)
    (16K,0.28)
    (18K,0.27)
};

\end{groupplot}
\end{tikzpicture}

\captionsetup{skip=4pt}
\caption{Comparison of Standard BPE and Pruned BPE for reducing encoded length}
\label{fig:standard-pruned-comparison}
\end{figure}
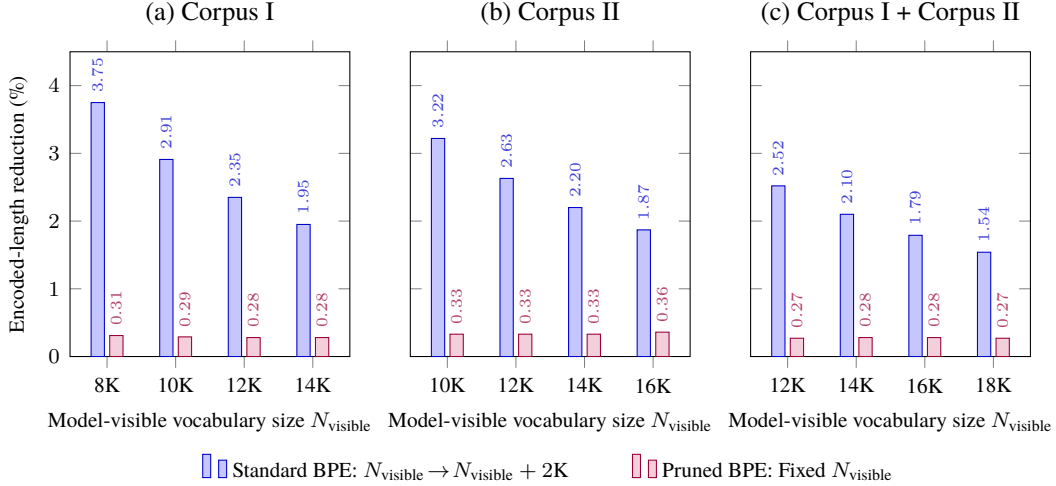

Figure~\ref{fig:standard-pruned-comparison} compares two different ways of reducing encoded length. Each panel corresponds to one training corpus: Corpus I, Corpus II, or the combined Corpus I + Corpus II. 
At each model-visible vocabulary size \(N_{\mathrm{visible}}\), the Standard BPE bar shows the marginal encoded-length reduction obtained by increasing the vocabulary from \(N_{\mathrm{visible}}\) to \(N_{\mathrm{visible}}+2\mathrm{K}\). In contrast, the Pruned BPE bar shows the additional reduction obtained relative to Standard BPE at the same model-visible vocabulary size \(N_{\mathrm{visible}}\), using a 40\% exposure threshold. Thus, the Standard BPE bars measure the benefit of adding 2K more model-visible tokens, whereas the Pruned BPE bars measure the benefit of reallocating the existing vocabulary capacity without increasing its size. Each tokenizer was evaluated on the same corpus used for its training.

Although the absolute encoded-length reduction achieved by Pruned BPE relative to Standard BPE may appear small, it should be interpreted in relation to the diminishing marginal gain obtained by expanding the Standard BPE vocabulary. Across the configurations shown in Figure~\ref{fig:standard-pruned-comparison}, increasing the Standard BPE vocabulary by 2K tokens reduces encoded length by approximately 1.5\%--3.8\%, while Pruned BPE provides an additional reduction of approximately 0.27\%--0.36\% under the same model-visible vocabulary budget. In later vocabulary ranges, where the marginal benefit of adding another 2K Standard BPE tokens is often only about 2\%, an additional reduction of roughly 0.3\% from pruning and token reallocation represents a meaningful fraction of that remaining gain. The figure therefore shows that Pruned BPE can improve encoding efficiency without requiring a larger model-visible vocabulary.

\subsection{Vocabulary-only Minimum-token Evaluation}
The preceding experiments use each tokenizer's native encoding procedure. Standard BPE applies learned merge rules according to their merge ranks, while Pruned BPE applies the retained merge rules and then recursively expands internal-only tokens. However, vocabulary construction and tokenizer inference are distinct components of a tokenization system, and different inference procedures may produce different segmentations from the same vocabulary \cite{uzan2024greed}. Schmidt et al.\ introduced PathPiece, which uses minimum-token segmentation for a given vocabulary \cite{schmidt2024tokenization}. Motivated by this line of work, an additional vocabulary-only evaluation was performed to determine whether the advantage of Pruned BPE persists when both visible vocabularies are encoded using the same minimum-token objective.

The Standard BPE and Pruned BPE tokenizers trained on Corpus I~+~Corpus II at model-visible vocabulary sizes of 12K, 14K, 16K, and 18K were evaluated using an independently implemented minimum-token dynamic-programming (DP) segmentation algorithm described below. The 40\% exposure-threshold configurations were used for Pruned BPE, consistent with the main comparison in Subsection~\ref{subsec:comp_std_bpe}. Each tokenizer was evaluated on Corpus I, Corpus II, and their combination.

For this evaluation, only the token strings in the corresponding model-visible \texttt{vocab.txt} file were made available to the encoder. Merge ranks, merge-tree child relationships, token-frequency statistics, and Pruned BPE's internal-only vocabulary were not used. In particular, tokens from \texttt{inter\_vocab.txt} were excluded because they are not part of the vocabulary exposed to the downstream model. The same pretokenization procedure described in Subsection~\ref{subsec:pretokenization} was applied before both native BPE encoding and minimum-token DP encoding. The optimization is therefore exact within each shared pretokenization boundary rather than across boundaries that neither tokenizer is permitted to cross.

For a pretokenized byte sequence \(x\) of length \(n\), let \(D(i)\) denote the minimum number of visible tokens required to encode the suffix beginning at byte position \(i\). The base case is \(D(n)=0\), and, for \(0 \leq i < n\), the recurrence is

\[
D(i)
=
\min_{\substack{
t \in V \\
x_{i:i+|t|}=t
}}
\left\{
1 + D\bigl(i+|t|\bigr)
\right\}.
\]

Here, \(V\) is the tokenizer's model-visible vocabulary and \(x_{i:i+|t|}\) is the byte sequence beginning at position \(i\) with the same length as token \(t\). Because all 256 individual byte tokens are visible, at least one valid transition is available at every position. The values are computed from right to left, and a trie is used to enumerate the vocabulary tokens beginning at each position without scanning the complete vocabulary. If \(\ell_{\max}\) is the maximum visible-token length, the encoding requires \(O(n\ell_{\max})\) time and \(O(n)\) DP storage, excluding the vocabulary trie.

Let \(L_{\mathrm{std}}^{\mathrm{DP}}(N)\) and
\(L_{\mathrm{pruned},40}^{\mathrm{DP}}(N)\) denote the encoded lengths produced by this shared DP encoder using, respectively, the Standard BPE and 40\%-threshold Pruned BPE visible vocabularies of size \(N\). The Pruned BPE reduction under the shared DP encoder is defined as

\[
\Delta_{\mathrm{DP}}(N)
=
\frac{
    L_{\mathrm{std}}^{\mathrm{DP}}(N)
    - L_{\mathrm{pruned},40}^{\mathrm{DP}}(N)
}{
    L_{\mathrm{std}}^{\mathrm{DP}}(N)
}
\times 100\%.
\]

\begin{figure}[!htbp]
\centering

\begin{tikzpicture}
\begin{groupplot}[
    group style={
        group size=3 by 1,
        horizontal sep=0.8cm,
        ylabels at=edge left,
        yticklabels at=edge left
    },
    width=0.32\textwidth,
    height=0.34\textwidth,
    ybar,
    /pgf/bar width=5pt,
    ymin=0,
    ymax=0.4,
    ytick={0,0.1,0.2,0.3,0.4},
    ylabel={Encoded-length reduction (\%)},
    xlabel={Model-visible vocabulary size \(N_{\mathrm{visible}}\)},
    symbolic x coords={12K,14K,16K,18K},
    xtick=data,
    enlarge x limits=0.18,
    tick label style={font=\scriptsize},
    label style={font=\scriptsize},
    title style={font=\small},
    legend style={
        font=\scriptsize,
        draw=none,
        at={(0.5,-0.30)},
        anchor=north,
        legend columns=2,
        legend cell align={left},
        /tikz/every even column/.append style={column sep=2em}
    },
    nodes near coords,
    nodes near coords style={
        font=\tiny,
        rotate=90,
        anchor=west,
        /pgf/number format/fixed,
        /pgf/number format/precision=2,
        /pgf/number format/fixed zerofill
    },
    every axis plot/.append style={
        fill opacity=0.75,
        draw opacity=1
    }
]

\nextgroupplot[
    title={(a) Corpus I},
    symbolic x coords={12K,14K,16K,18K},
    xtick=data
]

\addplot[standardbpe] coordinates {
    (12K,0.27)
    (14K,0.26)
    (16K,0.24)
    (18K,0.22)
};

\addplot[prunedbpe] coordinates {
    (12K,0.29)
    (14K,0.26)
    (16K,0.25)
    (18K,0.23)
};

\nextgroupplot[
    title={(b) Corpus II},
    symbolic x coords={12K,14K,16K,18K},
    xtick=data
]

\addplot[standardbpe] coordinates {
    (12K,0.27)
    (14K,0.29)
    (16K,0.30)
    (18K,0.30)
};

\addplot[prunedbpe] coordinates {
    (12K,0.29)
    (14K,0.30)
    (16K,0.31)
    (18K,0.31)
};

\legend{
    Native BPE encoding,
    Minimum-token DP encoding
}

\nextgroupplot[
    title={(c) Corpus I + Corpus II},
    symbolic x coords={12K,14K,16K,18K},
    xtick=data
]

\addplot[standardbpe] coordinates {
    (12K,0.27)
    (14K,0.28)
    (16K,0.28)
    (18K,0.27)
};

\addplot[prunedbpe] coordinates {
    (12K,0.29)
    (14K,0.29)
    (16K,0.29)
    (18K,0.28)
};

\end{groupplot}
\end{tikzpicture}

\captionsetup{skip=4pt}
\caption{Comparison of native BPE encoding and vocabulary-only minimum-token DP encoding}
\label{fig:min-token-dp-comparison}
\end{figure}
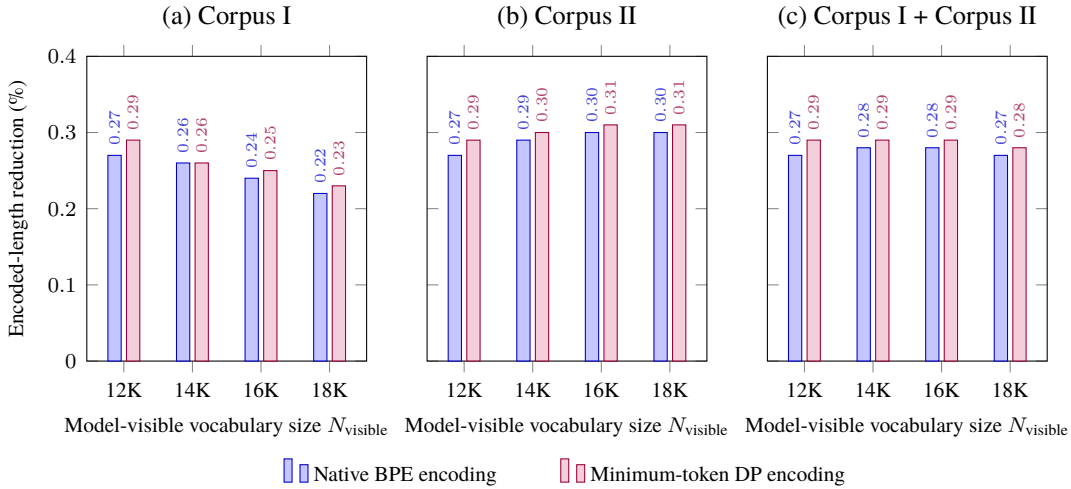

Figure~\ref{fig:min-token-dp-comparison} compares this reduction with the reduction obtained using the tokenizers' native BPE-based encoding procedures.

The minimum-token DP encoder produced fewer tokens than the corresponding native encoder for every Standard BPE and Pruned BPE vocabulary. Because this comparison is not directly related to the proposed pruning method, the detailed reductions are not reported here.

More importantly, Pruned BPE continued to produce shorter sequences in all 12 comparisons under the shared minimum-token objective. On the combined evaluation corpus, the reduction relative to Standard BPE was approximately 0.28\%--0.29\%. On Corpus I, it ranged from approximately 0.23\% to 0.29\%, while on Corpus II it ranged from approximately 0.29\% to 0.31\%. Across all configurations, the DP-based reduction ranged from approximately 0.23\% to 0.31\%, compared with approximately 0.22\% to 0.30\% under native BPE encoding.

Because both visible vocabularies were evaluated using the same segmentation objective, and because the DP encoder did not use merge ranks or internal-only merge nodes, the remaining improvement cannot be attributed solely to Standard BPE's merge-order encoding procedure. Instead, the result provides evidence that visibility pruning and token reallocation improve the composition of the model-visible vocabulary itself. This evaluation does not establish independence from every possible encoding algorithm, but it shows that the improvement persists when native merge-order effects are removed through the shared exact minimum-token DP encoder.

\subsection{Qualitative Token Analysis}
The Pruned BPE tokenizer trained on the combined corpus with an 18K target vocabulary and a 40\% exposure threshold was selected for qualitative inspection because it contains the largest model-visible vocabulary and a large set of internal-only tokens among the evaluated configurations. It therefore provides the broadest range of examples for examining the structural roles of internal-only tokens. All 419 internal-only tokens in this configuration remain direct or indirect components of model-visible tokens. Thus, although these tokens are not exposed to the language model, they are retained as functional nodes in the merge structure.

Many English examples are intermediate word fragments that are largely absorbed into longer visible tokens. For example, \enquote{\texttt{ecause}} is internal-only but is used directly to construct the visible tokens \enquote{\texttt{Because}}, \enquote{\texttt{ Because}}, and \enquote{\texttt{because}}. Similarly, \enquote{\texttt{vironment}} supports \enquote{\texttt{ environment}} and \enquote{\texttt{ Environment}}. Some internal fragments are shared by several different words: \enquote{\texttt{artment}} contributes to \enquote{\texttt{Department}}, \enquote{\texttt{department}}, and \enquote{\texttt{apartment}}, while \enquote{\texttt{ording}} contributes to both \enquote{\texttt{according}} and \enquote{\texttt{recording}}. These examples show that an internal-only token may remain useful as a reusable merge component even when exposing it as an independent model token provides limited benefit.

The Chinese tokens exhibit similar behavior at both the character and multi-character levels. For example, the internal token \cjktok{人民共和} is extended into the internal tokens \cjktok{人民共和国} and \cjktok{人民共和國}, which in turn support the visible tokens \cjktok{中华人民共和国} and \cjktok{中華人民共和國}, respectively. Other examples include \cjktok{西班} in \cjktok{西班牙}, \cjktok{澳大利} in \cjktok{澳大利亚}, and \cjktok{俱乐} in \cjktok{俱乐部}. Even complete characters may be internal-only when they occur primarily as components of longer visible tokens. For example, both \cjktok{尴} and \cjktok{尬} are internal-only components of the visible token \cjktok{尴尬}. A multi-level example is provided by \cjktok{葡萄牙}: the internal-only tokens \cjktok{葡} and \cjktok{萄} are first combined to form the visible token \cjktok{葡萄}, which is then used to construct the longer visible token \cjktok{葡萄牙}. This example also illustrates that a token may remain model-visible while serving as an intermediate component of another visible token.

Because the tokenizer operates on bytes, some internal-only tokens correspond to partial UTF-8 sequences rather than independently decodable text. For example, the internal byte sequence \enquote{\texttt{83 BD}} is shared by the visible Chinese characters \cjktok{能}, encoded as \enquote{\texttt{E8 83 BD}}, and \cjktok{都}, encoded as \enquote{\texttt{E9 83 BD}}. Similarly, the byte prefix \enquote{\texttt{E9 BC}} is used in visible tokens representing several different Chinese characters, including \cjktok{鼓}, \cjktok{鼠}, \cjktok{鼻}, and \cjktok{鼎}. The byte prefix \enquote{\texttt{E2 98}} likewise supports the black-star and white-star Unicode symbols, represented by \enquote{\texttt{U+2605}} and \enquote{\texttt{U+2606}}, respectively. Such fragments are necessary for preserving the learned byte-level merge structure but are poor candidates for occupying independent model-visible vocabulary slots.

A smaller number of internal tokens reflect structured web text. For example, \enquote{\texttt{ htt}} supports both \enquote{\texttt{ http}} and \enquote{\texttt{ https}}, while \enquote{\texttt{ www}} supports the visible token \enquote{\texttt{ www.}}; the leading spaces and the final period shown inside the quotation marks are part of the corresponding tokens. These examples indicate that visibility pruning is not limited to ordinary word morphology; it can also reclaim model-visible vocabulary slots occupied by intermediate fragments arising from multilingual byte sequences and structured text.

Not all internal-only tokens are obviously uninformative fragments. One example is \texttt{ournal}, which was also noted in the Introduction. It was not classified as an internal-only token by the tokenizer examined above, but it was classified as internal-only by the tokenizer trained on Corpus I with a 14K target vocabulary and a 40\% exposure threshold, even though it is a recognizable component of words such as \texttt{journal}, \texttt{journalism}, and \texttt{journalist}. This does not necessarily indicate an incorrect pruning decision: once longer tokens containing this fragment have been learned, \texttt{ournal} may have little final exposure as an independently emitted token. Moreover, its classification is configuration-dependent; a token that is internal-only for one training corpus, target vocabulary size, or exposure threshold may remain model-visible under another configuration. This example highlights that visibility pruning is determined by empirical final exposure rather than by whether a token appears linguistically meaningful in isolation.

\section{Discussion and Future Work}
Pruned BPE introduces several tradeoffs and limitations. It requires additional tokenizer training to produce enough eligible replacement tokens, and its effectiveness depends on the selected exposure threshold and on how well the tokenizer-training corpus represents downstream text. A threshold that is too high may cause useful tokens to be decomposed more often than the replacement candidates can compensate for. In addition, the present experiments evaluate tokenizer-level encoded length only. They do not yet establish whether Pruned BPE improves language-model training efficiency, validation loss, downstream task performance, or generation quality.

The vocabulary-only minimum-token evaluation provides additional evidence about the source of the encoded-length improvement. When the Standard BPE and Pruned BPE visible vocabularies were evaluated under the same exact minimum-token objective, Pruned BPE continued to produce shorter sequences in every tested configuration. Because this evaluation did not use native merge ranks or Pruned BPE's internal-only merge nodes, the result provides evidence that visibility pruning and token reallocation improve the composition of the model-visible vocabulary itself, rather than producing an advantage solely through the native encoding procedure. The minimum-token DP encoder was used as a diagnostic evaluation rather than as the proposed deployment encoder. The experiment therefore does not establish superiority under every possible inference method, nor does it address tokenizer runtime, language-model quality, or downstream training efficiency.

The most important direction for future work is controlled downstream language-model evaluation. Models with the same architecture, training data, optimization settings, and model-visible vocabulary size could be trained using Standard BPE and Pruned BPE tokenizers. Such experiments could compare validation loss, convergence behavior, training throughput, memory use, and downstream task performance. They could also test whether replacing rarely emitted visible tokens with better-exposed tokens leads to more effectively trained embeddings or a more balanced distribution of token usage.

Future work should also investigate automatic threshold selection and broader evaluation settings. Instead of selecting a fixed exposure ratio manually, the threshold could be chosen using held-out data or by estimating the tradeoff between the expansion cost of making a token internal-only and the compression benefit of the replacement token that occupies its visible slot. Additional experiments should cover larger multilingual corpora, code-focused datasets, and wider vocabulary ranges. Because visibility decisions are made after the merge structure has been trained, another possible direction is to export different domain-specific model-visible vocabularies from the same underlying BPE merge tree.

\section{Conclusion}
This paper introduced Pruned BPE, a post-training visibility-pruning and token-reallocation method that separates the BPE merge structure from the model-visible vocabulary. Low-exposure learned tokens remain available as internal merge nodes, while their visible vocabulary positions are reassigned to better-exposed candidates obtained through resumed Standard BPE training. Experiments across two corpora and their combination showed consistent encoded-length reductions relative to Standard BPE under the same model-visible vocabulary budget. The improvement also persisted when both visible vocabularies were evaluated under a shared exact minimum-token objective, providing evidence that the gain arises from more efficient model-visible vocabulary composition rather than solely from native BPE merge-order encoding. Although downstream language-model evaluation remains necessary, the results demonstrate that final exposure provides a practical basis for improving the allocation efficiency of a fixed BPE vocabulary.

\end{document}